\documentclass{article} 
\usepackage{iclr2026_conference,times}

\usepackage{amsmath,amsfonts,bm}

\def\eqref#1{equation~\ref{#1}}

\def\1{\bm{1}}

\DeclareMathAlphabet{\mathsfit}{\encodingdefault}{\sfdefault}{m}{sl}
\SetMathAlphabet{\mathsfit}{bold}{\encodingdefault}{\sfdefault}{bx}{n}

\usepackage{algorithm}
\usepackage{algpseudocode}
\usepackage{amsmath} 
\usepackage{xcolor}  
\usepackage{hyperref}
\usepackage{url}
\usepackage{booktabs}
\usepackage{graphicx}
\usepackage{enumitem}
\usepackage{bbm}
\usepackage[table]{xcolor} 

\title{ProtoTTA: \\Prototype-Guided Test-Time Adaptation}

\author{
\normalsize
\scalebox{0.92}{%
\begin{tabular}{c}
\textbf{Mohammad Mahdi Abootorabi\textsuperscript{1,3}, 
Parvin Mousavi\textsuperscript{2,3}, 
Purang Abolmaesumi\textsuperscript{1}, 
Evan Shelhamer\textsuperscript{1,3}} \\
[0.5em]
\textsuperscript{1} University of British Columbia \quad
\textsuperscript{2} Queen’s University \quad
\textsuperscript{3} Vector Institute
\end{tabular}
}
}

\iclrfinalcopy 
\begin{document}

\maketitle

\begin{abstract}
Deep networks that rely on prototypes—interpretable representations that can be related to the model input—have gained significant attention for balancing high accuracy with inherent interpretability, which makes them suitable for critical domains such as healthcare.
However, these models are limited by their reliance on training data, which hampers their robustness to distribution shifts.
While test-time adaptation (TTA) improves the robustness of deep networks by updating parameters and statistics, the prototypes of interpretable models have not been explored for this purpose.
We introduce \textit{\textbf{ProtoTTA}}, a general framework for prototypical models that leverages intermediate prototype signals rather than relying solely on model outputs.
\textit{ProtoTTA} minimizes the entropy of the prototype‑similarity distribution to encourage more confident and prototype-specific activations on shifted data.
To maintain stability, we employ geometric filtering to restrict updates to samples with reliable prototype activations, regularized by prototype-importance weights and model-confidence scores.
Experiments across four prototypical backbones on four diverse 
benchmarks spanning fine-grained vision, histopathology, and NLP demonstrate that ProtoTTA improves robustness over standard output entropy minimization while restoring correct semantic focus in prototype activations. 
We also introduce novel interpretability metrics and a vision-language model (VLM) evaluation framework to explain TTA dynamics, confirming that ProtoTTA restores human-aligned semantic focus and correlates reliably with VLM-rated reasoning quality.
Code is available at: \url{https://github.com/DeepRCL/ProtoTTA}.

\end{abstract}
\vspace{-1.4em}
\section{Introduction and Related Work}\label{sec::intro}
\vspace{-0.8em}
Prototype-based neural networks have emerged as a compelling solution for deep learning in critical domains, balancing high accuracy with inherent interpretability. Unlike standard black-box models, these architectures provide internal explainability by classifying inputs through similarity matching with learned prototypes, enabling intuitive ``this looks like that" reasoning with direct visual evidence for predictions \citep{chen2019looks}. This transparency has driven growing adoption in high-stakes applications, particularly healthcare \citep{wei2024mprotonet, vaseli2023protoasnet, sethi2025protoecgnet}, where understanding model decisions is as important as their accuracy.
The foundational model, ProtoPNet \citep{chen2019looks}, introduced the concept of classifying images by comparing input feature patches to learned prototypes. However, ProtoPNet relies on spatially rigid prototypes, limiting its ability to capture geometric variations. To address this, Deformable ProtoPNet \citep{donnelly2022deformable} replaced single rigid prototypes with flexible sub-prototypes that adapt their positions to better match input features. Subsequently, ProtoViT \citep{ma2024interpretable} extended this paradigm to Vision Transformer (ViT) \citep{dosovitskiy2020image} backbones with coherence-aligned sub-prototypes for capturing complex geometric variations. The field continues to evolve with innovations such as Hyperbolic Hierarchical Part Prototypes \citep{vaseli2025happi}, demonstrating sustained community interest in advancing prototype-based architectures.

\begin{figure}[t!]
\centering
  \includegraphics[width=0.86\textwidth]{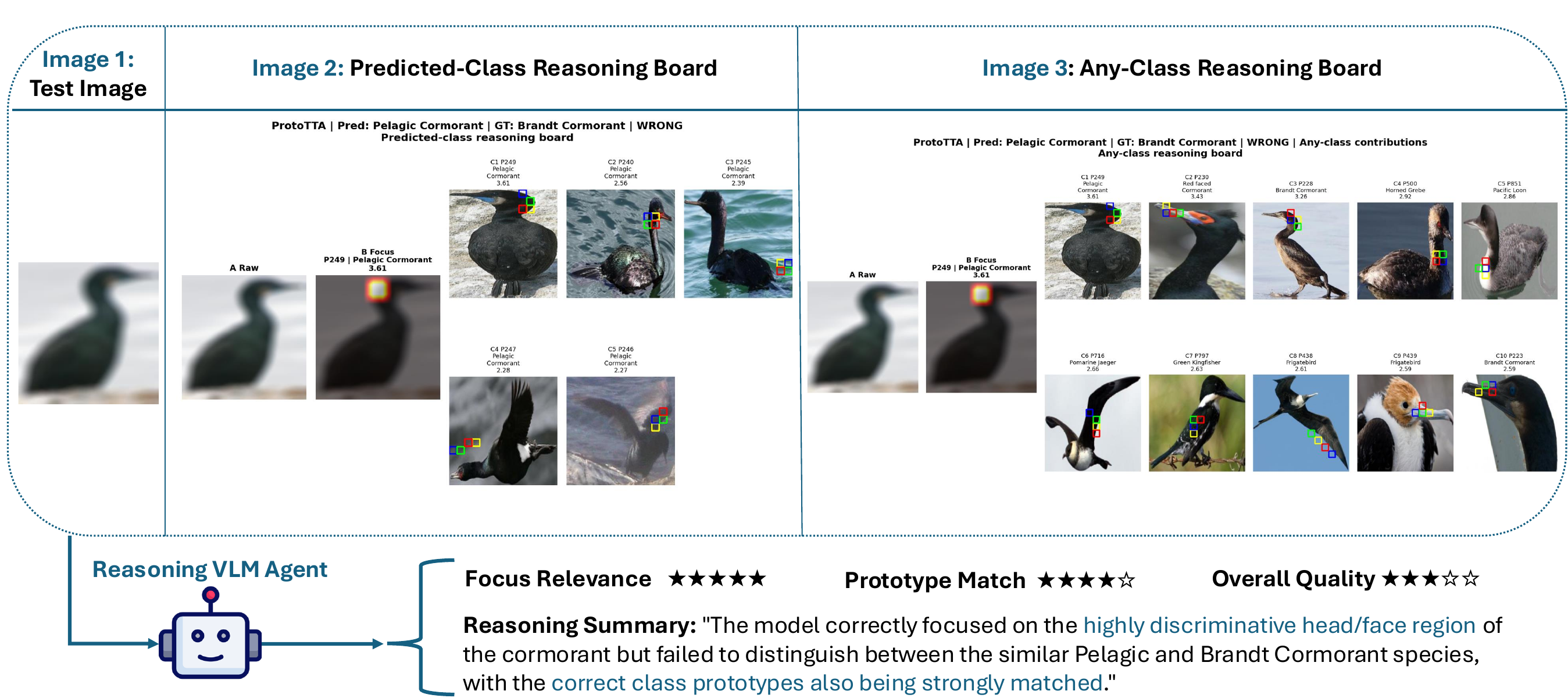}
\caption{
The VLM agent scores prototype-based reasoning boards, providing language-grounded evaluation of TTA dynamics unavailable for black-box methods.
}
  \label{fig:vlm}
\vspace{-1.6em}
\end{figure}

\vspace{-0.3em}
Despite these architectural advances, prototype-based models remain vulnerable to distribution shifts, where the discrepancy between training and test distributions degrades the semantic validity of selected prototypes. To address distribution shifts in general deep learning, Test-Time Adaptation (TTA) \citep{Liang_2024} has emerged as a paradigm to dynamically adjust models at inference using unlabeled target-domain data. Foundational methods like Tent \citep{wang2020tent} update normalization parameters via entropy minimization, while subsequent approaches have improved stability and efficiency through active sample selection (EATA) \citep{niu2022efficient}, sharpness-aware minimization (SAR) \citep{niu2023towards}, or enforcing consistency across augmented views (MEMO) \citep{zhang2022memo}. 
Beyond standard discriminative tasks, TTA has expanded to generative models \citep{prabhudesai2023diffusion} and even LLM agents that adapt to unseen environments by adjusting to local observation formats \citep{chen2025grounded}.
However, the intersection of TTA and model interpretability remains largely unexplored. Existing TTA methods treat underlying networks as black boxes, operating only on output logits or global feature statistics. While recent work such as TIDE \citep{agarwal2025tide} shows that training with local concept supervision can enable test-time correction, it requires specialized training pipelines and externally generated concept annotations. In the context of prototype-based models, current approaches overlook the rich interpretable intermediate signals, such as prototype activation patterns, spatial similarity maps, and semantic feature associations, that distinguish these architectures from conventional networks. This raises a fundamental question: Can we leverage a model's inherent interpretability signals to enable more semantic and reliable test-time adaptation? 
To this end, we introduce \textit{ProtoTTA}, the first TTA framework designed specifically for prototype-based architectures. Beyond improving robustness, our framework provides diagnostic transparency: we can trace why adaptation succeeds or fails by observing whether the model reactivates domain-invariant prototypes suppressed by corruption. To validate this, we introduce interpretability-aware metrics that assess semantic consistency, prototype alignment, and prediction stability, complementing standard accuracy measures with insights into the model's reasoning process. We further introduce a vision-language model (VLM)-based evaluation framework that explains TTA dynamics through language and prototype evidence quantitatively and qualitatively (Figure \ref{fig:vlm}), and show that our metrics correlate strongly with VLM-rated reasoning quality.

\begin{figure}[t!]
\centering
  \includegraphics[width=0.9\textwidth]{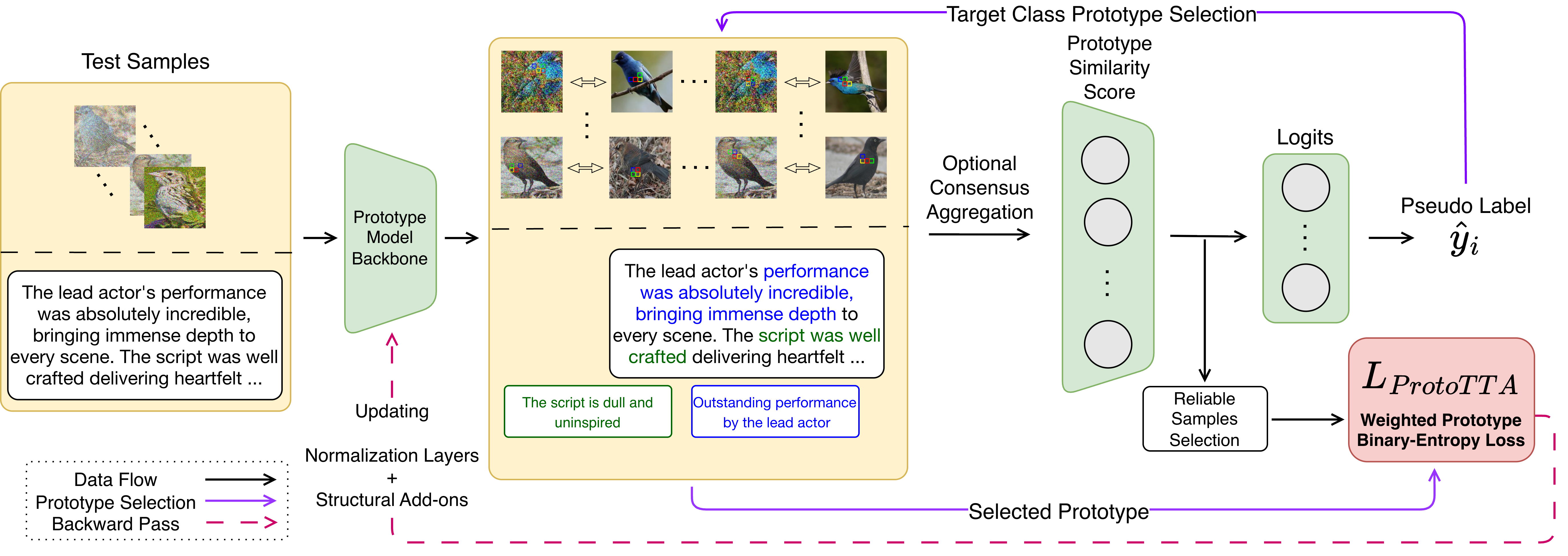}
\caption{Overview of the \textit{ProtoTTA} pipeline. The loss minimizes the binary entropy of prototype similarities for decisive activations. Geometric Filtering masks uncertain samples while Consensus Aggregation refines scores. Finally, updates to normalizations and structural add-ons (e.g., attention biases, $1\times1$ convolutions) restore semantic focus under domain shifts.}
  \label{fig:overview}
\vspace{-1.8em}
\end{figure}

\vspace{-1.3em}
\section{Method}
\vspace{-1.1em}
Prototype-based architectures provide three forms of granular interpretability absent in black-box models: (i) similarity scores quantifying input-prototype matching (prototype activations), (ii) prototype-to-class weights in the final classification head, and (iii) spatial localization binding prototypes to specific image regions. We leverage these intrinsic properties to design a specialized TTA method. Figure \ref{fig:overview} illustrates the ProtoTTA pipeline. The core intuition behind our approach is that distribution shifts can corrupt the prototype selection mechanism, causing the model to activate semantically irrelevant prototypes and suppress correct ones, degrading both accuracy and interpretability. Our adaptation strategy must therefore encourage reactivation of relevant prototypes while suppressing spurious activations. We achieve this by minimizing the entropy of prototype activations, compelling the model toward confident and unambiguous prototype matching.

\vspace{-0.5em}
To prevent adaptation on ambiguous or corrupted samples that could destabilize the model, we update only on a reliable set $\mathcal{R}$ using geometric filtering: we select samples whose maximum prototype similarity (after sub-prototype aggregation) exceeds a threshold $\tau$, optionally combined with a low-entropy constraint on the prediction distribution to ensure model confidence and sample utility. For each test sample $x_i \in \mathcal{R}$, the forward pass yields a pseudo-label $\hat{y}_i$, and we focus our adaptation specifically on the set of target prototypes $\mathcal{P}_t$ associated with this pseudo-label. 
Unlike standard TTA methods that minimize Shannon entropy on output logits, a distinct challenge here is that prototype activations are similarity scores (e.g., cosine similarity $\in [-1, 1]$) rather than probability distributions.
Moreover, unlike logits, where only one class should dominate, prototype activations represent independent matching signals where each prototype can legitimately exhibit strong or weak similarity. To address this, we map the raw similarity scores $s_{ip}$ into a probability space $\bar{s}_{ip} \in [0, 1]$ using linear scaling (for pre-normalized similarities) or other normalization schemes as needed, such as log-scaling. We then minimize the binary entropy of each mapped activation independently, which encourages each prototype to produce decisive similarities (near 0 or 1) rather than ambiguous mid-range values. In this mapped space, a value of $0.5$ corresponds to maximum entropy (equivalent to cosine similarity of 0), indicating high uncertainty or noise that should be suppressed. The adaptation objective is:

\vspace{-1.2em}
\begin{equation}
\mathcal{L}_{\text{ProtoTTA}} = \frac{1}{|\mathcal{R}|} \sum_{i \in \mathcal{R}} c_i \cdot \sum_{p \in \mathcal{P}_t} w_p \cdot H(\bar{s}_{ip})
\end{equation}
\vspace{-1.2em}

\noindent where $H(\bar{s}_{ip}) = - \bar{s}_{ip} \log(\bar{s}_{ip}) - (1 - \bar{s}_{ip}) \log(1 - \bar{s}_{ip})$ is the binary entropy of the mapped prototype activation. $c_i$ is the confidence score of sample $x_i$; $w_p$ represents the prototype weights from the final classification head for that specific class; and $\mathcal{P}_t$ is the set of target prototypes derived from pseudo-label $\hat{y}_i$. Minimizing this objective suppresses low-activation noise and reinforces strong prototype associations. Standard backbones with sub-prototypes typically derive the representative similarity score via maximum pooling or global averaging. However, maximum pooling is sensitive to outliers, while averaging weakens decisive signals by mixing them with low-activation noise. To mitigate this, we use a Top-K Mean strategy that aggregates the $k$ most relevant scores to enforce robust consensus. Finally, we update normalization parameters to correct global shifts and fine-tune structural add-ons, specifically attention biases (for Transformers) or $1\times1$ convolutions (for CNNs). These targeted updates recalibrate the model's semantic focus under domain shift while preserving the structural knowledge learned during training (Algorithm Pseudocode in Appendix \S\ref{sec:peseudocode}).

\begin{table*}[t!]
    \centering
    \caption{
    Test-time adaptation performance on CUB-200-C (Bird Classification) across diverse corruption types. We report accuracy (\%) for ProtoViT (Transformer-based). The final column shows the mean $\pm$ std across all corruptions. Best results in \textbf{bold}, second-best \underline{underlined}.
    }
    \label{tab:full_results}
    \vspace{0.05in}
    
    \resizebox{\textwidth}{!}{%
        \begin{tabular}{l|cccc>{\columncolor{gray!5}}c|cc>{\columncolor{gray!5}}c|ccc>{\columncolor{gray!5}}c|cccc>{\columncolor{gray!5}}c|c}
            \toprule
            & \multicolumn{5}{c|}{\textbf{Noise}} & \multicolumn{3}{c|}{\textbf{Blur}} & \multicolumn{4}{c|}{\textbf{Weather}} & \multicolumn{5}{c|}{\textbf{Digital}} & \\
            \textbf{Method} & Gauss & Shot & Impul & Speck & \textbf{Avg} & Defoc & Gauss & \textbf{Avg} & Brit & Fog & Frost & \textbf{Avg} & Contr & Elast & Jpeg & Pixel & \textbf{Avg} & \textbf{Total} \\
            \midrule
            
            \multicolumn{19}{c}{\textit{\textbf{Backbone: ProtoViT (LN) ~|~ Dataset: CUB-200-C}}} \\
            \midrule
            Unadapted & 37.3 & 36.5 & 39.6 & 48.8 & 40.5 & 42.8 & 39.0 & 40.9 & 69.2 & 61.6 & 43.4 & 58.1 & 49.6 & 71.0 & 67.7 & 68.1 & 64.1 & 51.9 \scriptsize{$\pm$ 13.0} \\
            Memo & 37.2 & 36.5 & 38.6 & 48.3 & 40.2 & 43.3 & 39.2 & 41.2 & 69.8 & 62.9 & 43.5 & 58.7 & 53.3 & 71.7 & \underline{68.2} & 69.8 & 65.8 & 52.5 \scriptsize{$\pm$ 13.5} \\
            SAR & 40.0 & 38.4 & 41.3 & 49.3 & 42.2 & 43.0 & 38.5 & 40.7 & 70.1 & 62.7 & 45.2 & 59.3 & 47.0 & 71.4 & 67.7 & 68.3 & 63.6 & 52.5 \scriptsize{$\pm$ 12.8} \\
            Tent & 45.8 & 41.2 & 40.2 & 45.5 & 43.2 & 41.5 & 39.2 & 40.4 & 72.2 & 63.0 & 49.9 & 61.7 & 52.9 & 73.8 & 67.0 & 70.3 & 66.0 & 54.0 \scriptsize{$\pm$ 12.8} \\
            EATA & \underline{50.1} & \underline{51.4} & \underline{51.7} & \underline{61.4} & \underline{53.7} & \underline{45.1} & \underline{41.5} & \underline{43.3} & \underline{73.4} & \underline{69.4} & \underline{52.7} & \underline{65.2} & \underline{55.3} & \underline{74.8} & 68.0 & \underline{70.8} & \underline{67.2} & \underline{58.9 \scriptsize{$\pm$ 10.8}} \\
            
            \rowcolor{gray!10} 
            \textbf{\textit{ProtoTTA} (Ours)} & \textbf{52.0} & \textbf{52.3} & \textbf{53.6} & \textbf{65.0} & \textbf{55.7} & \textbf{46.7} & \textbf{43.3} & \textbf{45.0} & \textbf{73.9} & \textbf{70.1} & \textbf{53.1} & \textbf{65.7} & \textbf{56.1} & \textbf{75.0} & \textbf{68.3} & \textbf{72.3} & \textbf{67.9} & \textbf{60.1 \scriptsize{$\pm$ 10.6}} \\
            
            \midrule

        \end{tabular}%
    }
    \vspace{-1.5em}
\end{table*}

\begin{table*}[t]\centering\caption{
Test-time adaptation on Amazon-C review classification using ProtoLens trained on Yelp. We report accuracy (\%) across 5 textual corruption types at 4 severity levels ($s \in \{20, 40, 60, 80\}$). The final column shows the mean $\pm$ std across all 20 scenarios. Best in \textbf{bold}, second-best \underline{underlined}.
}\label{tab:nlp_results}\vspace{0.05in}\resizebox{\textwidth}{!}{%
    \begin{tabular}{l|cccc>{\columncolor{gray!5}}c|cccc>{\columncolor{gray!5}}c|cccc>{\columncolor{gray!5}}c|cccc>{\columncolor{gray!5}}c|cccc>{\columncolor{gray!5}}c|c}
        \toprule
        & \multicolumn{5}{c|}{\textbf{Qwerty}} & \multicolumn{5}{c|}{\textbf{Swap}} & \multicolumn{5}{c|}{\textbf{Remove}} & \multicolumn{5}{c|}{\textbf{Mixed}} & \multicolumn{5}{c|}{\textbf{Aggressive}} & \\
        \textbf{Method} & 20 & 40 & 60 & 80 & \textbf{Avg} & 20 & 40 & 60 & 80 & \textbf{Avg} & 20 & 40 & 60 & 80 & \textbf{Avg} & 20 & 40 & 60 & 80 & \textbf{Avg} & 20 & 40 & 60 & 80 & \textbf{Avg} & \textbf{Total} \\
        \midrule
        
        Unadapted & \underline{89.4} & 85.4 & 80.0 & 71.5 & 81.6 & 89.9 & \underline{84.4} & 76.2 & \underline{70.6} & \underline{80.3} & \textbf{90.5} & 86.8 & \underline{82.5} & \textbf{76.8} & \underline{84.2} & \underline{84.1} & \underline{84.1} & \underline{84.1} & \underline{80.3} & \underline{83.2} & \underline{87.5} & \underline{79.6} & 71.2 & 61.3 & 74.9 & 80.81 \scriptsize{$\pm$ 7.43} \\
        
        SAR & 81.3 & 85.7 & 81.3 & 66.5 & 78.7 & 88.7 & 82.8 & 69.8 & 63.9 & 76.3 & 81.3 & \underline{87.1} & 80.3 & 70.8 & 79.8 & 81.3 & 82.8 & 81.3 & 77.4 & 80.7 & 81.3 & 76.6 & \textbf{78.1} & 55.2 & 72.8 & 77.65 \scriptsize{$\pm$ 8.19} \\
        
        Tent & 87.5 & 85.2 & \textbf{90.6} & 68.5 & \underline{82.9} & \textbf{90.2} & 84.1 & 75.4 & 68.9 & 79.6 & 87.5 & 86.8 & 81.4 & 75.1 & 82.7 & 81.3 & 83.6 & 81.3 & 79.0 & 81.3 & 81.3 & 77.9 & \textbf{78.1} & 58.4 & 73.9 & 80.08 \scriptsize{$\pm$ 7.80} \\
        
        EATA & 87.5 & 85.4 & \textbf{90.6} & \underline{71.6} & \textbf{83.8} & 89.8 & 84.2 & \underline{76.3} & \textbf{70.7} & 80.2 & 87.5 & 86.8 & 82.5 & \underline{76.8} & 83.4 & 81.3 & \underline{84.1} & 81.3 & 80.2 & 81.7 & 81.3 & 79.5 & \textbf{78.1} & \underline{61.6} & \underline{75.1} & \underline{80.84 \scriptsize{$\pm$ 6.91}} \\
        
        \midrule
        \rowcolor{gray!10} 
        \textbf{\textit{ProtoTTA} (Ours)} & \textbf{89.5} & \textbf{86.2} & 81.1 & \textbf{72.2} & 82.2 & \underline{90.0} & \textbf{85.0} & \textbf{76.4} & 69.9 & \textbf{80.3} & \underline{90.5} & \textbf{87.7} & \textbf{83.2} & 76.1 & \textbf{84.4} & \textbf{85.0} & \textbf{85.0} & \textbf{85.0} & \textbf{81.3} & \textbf{84.0} & \textbf{88.1} & \textbf{80.6} & \underline{71.9} & \textbf{62.3} & \textbf{75.7} & \textbf{81.33 \scriptsize{$\pm$ 7.45}} \\
        \bottomrule
    \end{tabular}%
}
    \vspace{-1.5em}
\end{table*}

\begin{table*}[t]
    \centering
    \caption{
    Efficiency and Interpretability Analysis across distinct backbones/datasets. 
    }
    \label{tab:efficiency_interpretability}    
    \resizebox{\textwidth}{!}{%
        \begin{tabular}{l|ccc|cc}
            \toprule
            & \multicolumn{3}{c|}{\textbf{Interpretability Metrics}} & \multicolumn{2}{c}{\textbf{Efficiency Metrics}} \\
            \textbf{Method} & \textbf{Semantic Consis. (PAC)} $\uparrow$ & \textbf{Proto. Alignment (PCA-W)} $\uparrow$ & \textbf{Prediction Stability} $\uparrow$ & \textbf{Selection Rate} $\downarrow$ & \textbf{Rel. Speed} $\uparrow$ \\
            \midrule
            
            \multicolumn{6}{c}{\textit{\textbf{Backbone: ProtoViT (LN + Attention Biases) ~|~ Dataset: CUB-200-C}}} \\
            \midrule
            Unadapted & $88.2 \pm 3.6$ & $70.8 \pm 11.2$ & $54.1 \pm 14.6\%$ & 0.0\% & 99.8\% \\
            Memo & $88.3 \pm 3.6$ & $71.1 \pm 11.2$ & $54.4 \pm 14.6\%$ & 100.0\% & 2.3\% \\
            SAR & $88.9 \pm 3.5$ & $74.3 \pm 10.3$ & $57.8 \pm 14.0\%$ & 98.9\% & 48.7\% \\
            Tent & $88.3 \pm 5.8$ & $59.1 \pm 32.3$ & $46.5 \pm 30.2\%$ & 100.0\% & 91.7\% \\
            EATA & \underline{$91.3 \pm 3.1$} & \underline{$81.1 \pm 7.7$} & \underline{$66.5 \pm 11.8\%$} & \underline{68.1\%} & \underline{94.9\%} \\
            
            \rowcolor{gray!10} 
            \textbf{\textit{ProtoTTA} (Ours)} & \boldmath{$91.9 \pm 2.9$} & \boldmath{$82.6 \pm 7.1$} & \textbf{68.7 $\pm$ 11.5\%} & \textbf{58.0\%} & \textbf{95.7\%} \\
            
            \midrule
            
            
            

            
            \multicolumn{6}{c}{\textit{\textbf{Backbone: ProtoLens (LN + Attention Biases) ~|~ Dataset: Amazon-C}}} \\
            \midrule
            Unadapted & $17.9 \pm 0.7$ & $64.4 \pm 3.9$ & $50.7 \pm 0.5\%$ & 0.0\% & 100.0\% \\
            SAR & \textbf{20.8 $\pm$ 6.2} & $63.8 \pm 3.8$ & \textbf{51.6 $\pm$ 4.7\%} & \underline{7.8\%} & 62.8\% \\
            Tent & $18.1 \pm 2.8$ & $64.1 \pm 4.1$ & $50.1 \pm 1.9\%$ & 100.0\% & 98.0\% \\
            EATA & \underline{$18.6 \pm 2.4$} & \underline{$64.5 \pm 3.6$} & $50.2 \pm 1.9\%$ & \textbf{1.4\%} & \underline{100.0\%} \\
            
            \rowcolor{gray!10}
            \textbf{\textit{ProtoTTA} (Ours)} & $18.2 \pm 0.7$ & \textbf{64.8 $\pm$ 3.9} & \underline{50.8 $\pm$ 0.4\%} & 28.2\% & \textbf{100.0\%} \\
            
            \bottomrule
        \end{tabular}%
    }
        \vspace{-0.9em}
\end{table*}

\vspace{-1.2em}
\section{Experiments and Results}
\vspace{-1.0em}
\paragraph{Dataset and Experimental Settings.} 
We evaluate ProtoTTA on vision (CUB-200-C \citep{wah_branson_welinder_perona_belongie_2011}) 
and NLP (Amazon-C) benchmarks using ProtoViT \citep{ma2024interpretable} and 
ProtoLens \citep{wei-zhu-2025-protolens} as primary backbones; results on 
ProtoPNet \citep{chen2019looks}/SICAPv2-C \citep{silvarodriguez2020sicapv2} and 
ProtoPFormer \citep{xue2024protopformer}/Stanford Dogs-C \citep{khosla2011novel} 
are in Appendix~\S\ref{sec:additional_backbones}.
Beyond classification accuracy, we report three interpretability-oriented metrics: \textit{(i) Prototype Activation Consistency (PAC)}, measuring cosine similarity between clean and adapted activations to quantify semantic stability; \textit{(ii) Weighted Prototype Alignment (PCA-W)}, evaluating if highly activated prototypes align with the ground truth, weighted by activation strength and classification-layer importance; and \textit{(iii) Prediction Stability}, measuring agreement between clean and adapted predictions. For efficiency, we additionally report the \textit{Selection Rate} (fraction of test samples triggering adaptation) and \textit{Relative Speed} compared to the unadapted model.
More details about experiments, datasets, and metrics can be found in Appendix \S\ref{sec:dataset}.

\vspace{-1.3em}
\paragraph{Results.}
\textit{ProtoTTA} consistently outperforms all baselines across vision and NLP benchmarks (Tables~\ref{tab:full_results},~\ref{tab:nlp_results}), demonstrating that prototype-specific signals enable more effective adaptation than treating interpretable models as black boxes. Critically, \textit{ProtoTTA} is entirely
source-free, whereas EATA, the closest competitor, requires ${\approx}2000$ source samples. ProtoViT's sub-prototype structure provides ample capacity for semantic refocusing, yielding substantial gains. We note that blur corruptions are uniquely challenging over all vision backbones, suggesting patch-prototype matching relies heavily on vulnerable high-frequency local features. Cross-domain NLP results confirm generalization beyond vision despite the added challenge of shared non-class-specific prototypes in ProtoLens. Generalization to CNN and additional Transformer backbones as well as integration with existing TTA methods is demonstrated in Appendix~\S\ref{sec:additional_backbones}. 
Beyond accuracy, Table~\ref{tab:efficiency_interpretability} reveals a crucial takeaway: \textit{ProtoTTA} achieves superior performance while adapting on fewer samples, maintaining both efficiency and interpretability. High PAC and PCA-W
scores confirm that adaptation restores the model's original reasoning process, reactivating correct-class prototypes suppressed by corruption, rather than merely
improving output statistics (qualitative analysis in Appendix~\S\ref{sec:qualitative_examples}). Ablation studies (Appendix~\S\ref{sec:ablation}) validate our design choices, and continual adaptation experiments confirm resistance
to catastrophic forgetting (Appendix~\S\ref{sec:continual}).

\vspace{-1.3em}
\paragraph{VLM-Based Interpretability Evaluation.}
Prototype-based TTA is inherently explainable: reasoning boards expose which prototypes drive each prediction, enabling automated language-grounded evaluation unavailable for black-box methods. We pass reasoning boards from a representative
CUB-200-C subset to a VLM agent that scores focus relevance, prototype match, and overall adaptation quality (protocol in Appendix~\S\ref{sec:vlm_scoring}). Table~\ref{tab:vlm_scores} shows that \textit{ProtoTTA} achieves the highest scores
across all dimensions, with the largest gains in focus relevance ($+0.12$ over EATA) and prototype match ($+0.20$ over unadapted), confirming that adaptation improves semantic reasoning quality beyond accuracy.  
Furthermore, sample-level PCA-W correlates positively with VLM-rated overall quality across all methods (Pearson $r{=}0.53$, Spearman $\rho{=}0.59$), validating that it captures genuine interpretability signals aligned with human-proxy evaluation.
Notably, this correlation strengthens significantly under \textit{ProtoTTA} ($r{=}0.68$), as our adaptation explicitly suppresses the noise-induced semantic hallucinations that otherwise decouple mathematical activations from visual reality (analysis in Appendix~\S\ref{sec:vlm_corr_analysis}).
Ultimately, this framework enables a direct comparison between unadapted and adapted predictions, providing a language-grounded analysis of exactly how adaptation alters semantic focus and showcasing the unique power of explainable TTA (details in Appendix~\S\ref{sec:vlm_qual}).


\bibliography{iclr2026_conference}
\bibliographystyle{iclr2026_conference}

\newpage
\appendix

\section{Dataset and Experimental Details} \label{sec:dataset}

\paragraph{Vision Benchmarks and Models.} 
We evaluate ProtoTTA on two challenging vision datasets that require fine-grained discrimination. For \textit{CUB-200} \citep{wah_branson_welinder_perona_belongie_2011}, a fine-grained bird classification dataset (200 classes), we use the ProtoViT \citep{ma2024interpretable} model with a DeiT-S/16 backbone. This model contains 2,000 prototypes (10 per class, with 4 sub-prototypes each) and achieves 85.4\% accuracy on clean test data. We construct the \textit{CUB-200-C} dataset using 13 of the corruption types with severity 5, following ImageNet-C \citep{hendrycks2019robustness}, categorized as follows:
\begin{itemize}
    \item \textbf{Noise:} Gaussian, Shot, Impulse, Speckle
    \item \textbf{Blur:} Gaussian, Defocus
    \item \textbf{Weather:} Fog, Frost, Brightness
    \item \textbf{Digital:} JPEG Compression, Contrast, Pixelate, Elastic Transform
\end{itemize}

For \textit{SICAPv2} \citep{silvarodriguez2020sicapv2}, a challenging histopathology dataset for prostate cancer grading that requires distinguishing subtle morphological differences between cancer grades, we employ the legacy ProtoPNet architecture \citep{chen2019looks} with a VGG19-BN \citep{simonyan2014very} backbone. This model is configured with 50 prototypes (10 per class) and achieves 63.4\% accuracy on clean test data. 
To bridge the gap between ProtoPNet's distance-based metric and the probabilistic requirements of our adaptation method, we apply a specific transformation (a form of Log-Inverse Distance Kernel) to the prototype activations. ProtoPNet natively outputs minimum squared Euclidean distances, $d_{min}$. We map these to a probability distribution by first computing a raw similarity score $s_{raw}$ via logarithmic activation:
\begin{equation}
    s_{raw} = \log\left(\frac{d_{min} + 1.0}{d_{min} + 10^{-4}}\right)
\end{equation}
which is subsequently normalized to the $[0, 1]$ range. We construct \textit{SICAPv2-C} following the same corruption protocol as CUB-200-C.

For \textit{Stanford Dogs} \citep{khosla2011novel}, a fine-grained dog breed classification dataset (120 classes), we use the ProtoPFormer \citep{xue2024protopformer} model with a DeiT-S/16 backbone. ProtoPFormer extends prototype learning to Vision Transformers through a \textit{token reservation} mechanism: at the last transformer block, only the top-k most-attended patch tokens are retained for prototype comparison, concentrating the model's attention on discriminative image regions. The model employs a dual-branch architecture with 1,800 prototypes in total,  1,200 local patch-level prototypes (10 per class), and 600 global CLS-token
prototypes (5 per class), each of dimension 384, with predictions formed as an equal-weight combination of both branches. It achieves 90.75\% accuracy on clean test data. We construct \textit{Stanford Dogs-C} using the same 13 corruption types at severity 5 following ImageNet-C \citep{hendrycks2019robustness}, with the same category groupings as CUB-200-C. Results are reported in Appendix~\S\ref{sec:additional_backbones}.

\paragraph{NLP Benchmark and Model.} 
We evaluate ProtoLens \citep{wei-zhu-2025-protolens} for review sentiment classification. We use a ProtoLens model with an \textit{all-mpnet-base-v2} \citep{song2020mpnet} backbone, configured with 50 shared prototypes. Each prototype represents a semantic concept and is associated with some representative sub-sentences extracted from the training set via a 5-word sliding window for interpretability. The model was trained on the Yelp dataset (94.0\% accuracy) and achieves 91.97\% accuracy on the clean Amazon test dataset. We create the \textit{Amazon-C} benchmark following WildNLP \citep{rychalska2019models} with 5 corruption types applied across 4 severity levels (20\%, 40\%, 60\%, and 80\% word-level corruption), resulting in 20 total experimental settings. The corruptions are categorized as follows:
\begin{itemize}
    \item \textbf{Keyboard:} QWERTY
    \item \textbf{Character:} Swap, Remove Char
    \item \textbf{Combined:} Mixed, Aggressive
\end{itemize}

During test-time adaptation, we apply a temperature-scaled sigmoid function to the prototype similarities to map them into the $[0, 1]$ range with an appropriate semantic spread:
\begin{equation}
    p_{i} = \sigma(\tau \cdot s_{i}) = \frac{1}{1 + e^{-\tau \cdot s_{i}}}
\end{equation}
where $s_i$ is the cosine similarity for prototype $i$, and $\tau$ is a temperature hyperparameter (typically set to 5.0) that controls the sharpness of the resulting probability distribution.

\paragraph{Optimization.} 
Optimization is performed using Adam \citep{kingma2017adammethodstochasticoptimization} with a learning rate of $10^{-3}$ and momentum $\beta_1=0.9$. We use a test batch size of 128 and perform a single gradient update per batch, though our method remains effective across varying batch sizes. All of our experiments are conducted in non-episodic settings. We observed the same trend in episodic settings as well, and we plan to incorporate these findings into future designs.

\subsection{Evaluation Metrics} \label{sec:metrics}
To rigorously assess both performance and interpretability preservation, we report the following metrics alongside standard classification accuracy.

\paragraph{Prototype Activation Consistency (PAC).} 
PAC quantifies the semantic stability of the model by measuring the cosine similarity between the prototype activation vectors of the clean ($\mathbf{a}_i^{\text{clean}}$) and adapted ($\mathbf{a}_i^{\text{adapted}}$) inputs for a given sample $i$:
\begin{equation}
\text{PAC} = \frac{1}{N} \sum_{i=1}^{N} \frac{\mathbf{a}_i^{\text{clean}} \cdot \mathbf{a}_i^{\text{adapted}}}{\|\mathbf{a}_i^{\text{clean}}\|_2 \|\mathbf{a}_i^{\text{adapted}}\|_2}
\end{equation}
where $N$ is the total number of test samples. A higher PAC score indicates that the adaptation process preserves the original semantic representation of the input.

\paragraph{Weighted Prototype Alignment (PCA-W).} 
To verify that the model attends to semantically correct features, we introduce PCA-W. For a sample $i$ with ground truth class $y_i$, we analyze the set of top-$k$ activated prototypes $\mathcal{T}_i$. We weight each prototype $p \in \mathcal{T}_i$ by its contribution to the true class, defined as $c_{i,p} = a_{i,p} \cdot |W_{y_i, p}|$, where $a_{i,p}$ is the activation strength and $W_{y_i, p}$ is the weight connecting prototype $p$ to class $y_i$:
\begin{equation}
\text{PCA-W} = \frac{1}{N} \sum_{i=1}^{N} \frac{\sum_{p \in \mathcal{T}_i} c_{i,p} \cdot \mathbbm{1}[\text{class}(p) = y_i]}{\sum_{p \in \mathcal{T}_i} c_{i,p}}
\end{equation}
Unlike simple accuracy, PCA-W confirms that the model is right for the right reasons—specifically, that the high-activation prototypes actually belong to the correct semantic category.

\paragraph{Prediction Stability.} 
We measure the stability of the decision boundary by calculating the prediction agreement between the clean and adapted models:
\begin{equation}
\text{Prediction Stability} = \frac{1}{N} \sum_{i=1}^{N} \mathbbm{1}[\hat{y}_i^{\text{adapted}} = \hat{y}_i^{\text{clean}}]
\end{equation}
High Prediction Stability indicates that the adaptation improves robustness without arbitrarily flipping predictions for samples that were already handled correctly by the clean model.

\paragraph{Efficiency.} 
We report \textit{Selection Rate}, defined as the percentage of test samples that trigger the adaptation update ($N_{\text{adapted}} / N_{\text{total}}$), and \textit{Relative Speed}, defined as the ratio of the inference throughput (samples/sec) of the adapted model to that of the unadapted baseline.

\section{CNN and Additional Transformer Backbone Results}
\label{sec:additional_backbones}
We evaluate ProtoTTA on two additional backbones to demonstrate 
generalization across architectures. For the CNN-based ProtoPNet on SICAPv2-C, the architecture presents distinct challenges, including a lack of sub-prototypes, limited depth, and lower separation of prototypes in the representation space. 
These limitations reduce the available adaptation headroom. To address this, we introduce \textit{ProtoTTA+}, a hybrid approach that leverages logit distributions. 
By combining our prototype-aware loss ($W{=}0.7$) with standard entropy minimization on logits ($W{=}0.3$), \textit{ProtoTTA+} bridges this gap and achieves best-in-class 
performance. This demonstrates that our prototype-guided approach is complementary and can be seamlessly integrated with existing TTA methods rather than simply replacing them. 

Furthermore, for ProtoPFormer on Stanford Dogs-C—which extends prototype learning to Vision Transformers via token reservation—\textit{ProtoTTA+} achieves the best overall 
accuracy (57.7\%). This confirms that our hybrid approach generalizes effectively across diverse prototype architectures and complex fine-grained recognition tasks. These results demonstrate that our approach is complementary and can be effectively integrated with existing TTA methods such as SAR or Tent, rather than simply replacing them.

\begin{table*}[h]
    \centering
    \caption{
    Test-time adaptation performance on SICAPv2-C (histopathology 
    prostate cancer grading) using ProtoPNet (CNN-based), and on 
    Stanford Dogs-C (fine-grained dog breed classification) using 
    ProtoPFormer (Transformer-based). The final column shows the 
    mean $\pm$ std across all corruptions. Best results in 
    \textbf{bold}, second-best \underline{underlined}.
    }
    \label{tab:appendix_accuracy}
    \vspace{0.05in}
    \resizebox{\textwidth}{!}{%
        \begin{tabular}{l|cccc>{\columncolor{gray!5}}c|cc>{\columncolor{gray!5}}c|ccc>{\columncolor{gray!5}}c|cccc>{\columncolor{gray!5}}c|c}
            \toprule
            & \multicolumn{5}{c|}{\textbf{Noise}} & \multicolumn{3}{c|}{\textbf{Blur}} & \multicolumn{4}{c|}{\textbf{Weather}} & \multicolumn{5}{c|}{\textbf{Digital}} & \\
            \textbf{Method} & Gauss & Shot & Impul & Speck & \textbf{Avg} & Defoc & Gauss & \textbf{Avg} & Brit & Fog & Frost & \textbf{Avg} & Contr & Elast & Jpeg & Pixel & \textbf{Avg} & \textbf{Total} \\
            \midrule
            \multicolumn{19}{c}{\textit{\textbf{Backbone: ProtoPNet (BN + $1{\times}1$ Convs) ~|~ Dataset: SICAPv2-C}}} \\
            \midrule
            Unadapted      & 16.4 & 11.1 & 13.1 & 11.5 & 13.0 & 36.9 & 34.4 & 35.7 & 31.9 & 30.3 & 31.3 & 31.2 & 30.9 & 57.6 & 49.4 & 55.5 & 48.4 & 31.6 \scriptsize{$\pm$ 15.2} \\
            Memo           & \underline{58.5} & \textbf{58.7} & 57.5 & \underline{57.1} & \textbf{58.0} & 51.1 & 54.1 & 52.6 & \textbf{59.7} & 50.1 & 59.5 & \underline{56.4} & \textbf{44.7} & 56.8 & 51.7 & 60.2 & 53.4 & 55.4 \scriptsize{$\pm$ 4.5} \\
            SAR            & 58.2 & \underline{58.3} & 57.9 & 55.8 & 57.6 & \textbf{54.2} & \textbf{56.1} & \textbf{55.2} & 57.6 & \textbf{51.6} & 60.1 & \underline{56.4} & 43.5 & 57.5 & 51.5 & 61.6 & 53.5 & \underline{55.7 \scriptsize{$\pm$ 4.5}} \\
            Tent           & 56.6 & 54.6 & \textbf{58.8} & 54.9 & 56.2 & 53.0 & \underline{55.7} & \underline{54.4} & 52.1 & 45.0 & 58.4 & 51.8 & 39.8 & \underline{60.1} & 50.9 & 60.2 & 52.8 & 53.8 \scriptsize{$\pm$ 5.7} \\
            EATA           & 58.3 & \underline{58.3} & 58.1 & 55.7 & 57.6 & \underline{53.4} & 55.0 & 54.2 & 58.4 & \underline{50.7} & \textbf{61.3} & \textbf{56.8} & 43.3 & 58.2 & 52.2 & 60.3 & 53.5 & 55.6 \scriptsize{$\pm$ 4.7} \\
            \rowcolor{gray!10}
            \textbf{\textit{ProtoTTA} (Ours)}  & \textbf{59.4} & 58.1 & 57.9 & 55.1 & 57.6 & 51.6 & 54.4 & 53.0 & 56.5 & 47.9 & \underline{60.6} & 55.0 & \underline{43.8} & 58.7 & \underline{52.2} & \underline{61.7} & \underline{54.1} & 55.2 \scriptsize{$\pm$ 5.0} \\
            \rowcolor{gray!10}
            \textbf{\textit{ProtoTTA+} (Ours)} & 58.3 & 57.6 & \underline{58.6} & \textbf{57.3} & \textbf{58.0} & 51.9 & \textbf{56.1} & 54.0 & \underline{59.2} & 49.2 & 59.2 & 55.9 & 43.7 & \textbf{60.3} & \textbf{52.3} & \textbf{61.9} & \textbf{54.6} & \textbf{55.8 \scriptsize{$\pm$ 4.9}} \\
            \midrule
            \multicolumn{19}{c}{\textit{\textbf{Backbone: ProtoPFormer (LN + Attention Biases) ~|~ Dataset: Stanford Dogs-C}}} \\
            \midrule
            Unadapted & 46.6 & 46.2 & 48.1 & 55.6 & 49.1 & 40.4 & 37.4 & 38.9 & \textbf{69.6} & 51.2 & 48.0 & 56.3 & 46.5 & 57.8 & \textbf{68.5} & \textbf{70.8} & 60.9 & 52.8 \scriptsize{$\pm$ 10.6} \\
            Tent      & \underline{52.7} & 51.8 & \textbf{54.7} & \underline{60.7} & \underline{55.0} & \underline{45.8} & 42.3 & 44.1 & \underline{68.3} & \underline{60.4} & \underline{53.8} & \underline{60.8} & \underline{49.3} & 64.9 & 67.1 & 70.6 & \underline{63.0} & 57.1 \scriptsize{$\pm$ 8.6} \\
            SAR       & 36.7 & 40.9 & 41.8 & 48.1 & 41.9 & 31.5 & 30.0 & 30.8 & 59.6 & 34.4 & 30.6 & 41.5 & 31.8 & 50.3 & 59.5 & 63.1 & 51.2 & 42.9 \scriptsize{$\pm$ 11.5} \\
            EATA      & 52.0 & \underline{53.3} & 53.4 & 60.5 & 54.8 & \textbf{46.7} & \underline{43.7} & \textbf{45.2} & 68.2 & 59.7 & 52.9 & 60.3 & \textbf{49.5} & \underline{65.3} & 67.2 & \underline{70.7} & \textbf{63.2} & \underline{57.2 \scriptsize{$\pm$ 8.4}} \\
            \rowcolor{gray!10}
            \textbf{\textit{ProtoTTA+} (Ours)} & \textbf{53.8} & \textbf{54.9} & \underline{53.8} & \textbf{60.9} & \textbf{55.9} & 45.2 & \textbf{45.2} & \textbf{45.2} & 68.1 & \textbf{61.8} & \textbf{54.2} & \textbf{61.4} & 48.0 & \textbf{65.6} & \underline{67.9} & 70.0 & 62.9 & \textbf{57.7 \scriptsize{$\pm$ 8.4}} \\
            \bottomrule
        \end{tabular}%
    }
\end{table*}

\begin{table*}[t]
    \centering
    \caption{
    Efficiency and interpretability analysis for ProtoPNet on 
    SICAPv2-C and ProtoPFormer on Stanford Dogs-C. Best in 
    \textbf{bold}, second-best \underline{underlined}.
    }
    \label{tab:appendix_interp}
    \resizebox{\textwidth}{!}{%
        \begin{tabular}{l|ccc|cc}
            \toprule
            & \multicolumn{3}{c|}{\textbf{Interpretability Metrics}} 
            & \multicolumn{2}{c}{\textbf{Efficiency Metrics}} \\
            \textbf{Method} 
            & \textbf{Semantic Consis. (PAC)} $\uparrow$ 
            & \textbf{Proto. Alignment (PCA-W)} $\uparrow$ 
            & \textbf{Prediction Stability} $\uparrow$ 
            & \textbf{Selection Rate} $\downarrow$ 
            & \textbf{Rel. Speed} $\uparrow$ \\
            \midrule
            \multicolumn{6}{c}{\textit{\textbf{Backbone: ProtoPNet (BN + $1{\times}1$ Convs) ~|~ Dataset: SICAPv2-C}}} \\
            \midrule
            Unadapted          & \textbf{88.8 $\pm$ 2.6} & $51.3 \pm 5.9$ & $20.8 \pm 13.0\%$ & 0.0\% & 99.9\% \\
            Memo               & $86.1 \pm 0.4$ & $60.7 \pm 1.5$ & \textbf{31.1 $\pm$ 1.3\%} & 100.0\% & 45.4\% \\
            SAR                & $85.9 \pm 0.3$ & \textbf{62.0 $\pm$ 1.9} & \underline{28.3 $\pm$ 1.8\%} & 100.0\% & 22.6\% \\
            Tent               & $84.5 \pm 0.5$ & $60.9 \pm 2.6$ & $22.7 \pm 1.2\%$ & 100.0\% & 45.3\% \\
            EATA               & $86.2 \pm 0.4$ & \underline{61.9 $\pm$ 2.0} & $26.6 \pm 1.2\%$ & \underline{3.0\%} & \textbf{89.5\%} \\
            \rowcolor{gray!10}
            \textbf{\textit{ProtoTTA} (Ours)}  & \underline{87.7 $\pm$ 1.1} & $60.4 \pm 2.4$ & $27.9 \pm 2.1\%$ & \textbf{1.2\%} & \underline{75.3\%} \\
            \rowcolor{gray!10}
            \textbf{\textit{ProtoTTA+} (Ours)} & $87.4 \pm 0.6$ & $59.8 \pm 1.4$ & $27.2 \pm 2.2\%$ & 76.3\% & 45.1\% \\
            \midrule
            \multicolumn{6}{c}{\textit{\textbf{Backbone: ProtoPFormer (LN + Attention Biases) ~|~ Dataset: Stanford Dogs-C}}} \\
            \midrule
            Unadapted          & \underline{92.1} $\pm$ 0.4 & 53.5 $\pm$ 10.2 & 55.5\% & 0.0\%   & 100.0\% \\
            Tent               & 91.2 $\pm$ 0.2 & \underline{64.8} $\pm$ 7.3  & \underline{61.8\%} & 100.0\%  & 74.6\%  \\
            EATA               & 91.3 $\pm$ 0.3 & \textbf{65.1} $\pm$ 7.1  & \textbf{63.4\%} & \underline{69.7\%}  & 72.1\%  \\
            SAR                & \textbf{93.8} $\pm$ 0.2 & 35.3 $\pm$ 11.3 & 58.0\% & \textbf{0.7\%} & \textbf{94.5\%} \\
            \rowcolor{gray!10}
            \textbf{\textit{ProtoTTA+} (Ours)} & 91.2 $\pm$ 0.2 & 64.4 $\pm$ 7.4 & 58.6\% & 71.4\%  & \underline{82.5\%}  \\
            \bottomrule
        \end{tabular}%
    }
\end{table*}

\section{VLM-Based Interpretability Evaluation}
\label{sec:vlm_scoring}

\subsection{Scoring Protocol}
\label{sec:vlm_scoring_protocol}
We construct a fixed 100-sample subset of CUB-200-C by sampling 
uniformly across corruption types and severity levels. For each 
sample and each TTA method, we assemble a reasoning board comprising: (i) the corrupted test image, (ii) the predicted-class reasoning board showing the top-matched prototypes with their spatial focus regions highlighted, and (iii) the any-class contribution board showing the highest-contributing prototypes across all classes alongside their contribution scores. These boards are passed to a VLM agent (Qwen/Qwen3-VL-32B-Thinking \citep{bai2025qwen3vltechnicalreport}) that returns three scores on a 1--5 integer scale:
\begin{itemize}[leftmargin=1.5em,itemsep=0pt,topsep=2pt]
    \item \textbf{Focus Relevance:} whether the highlighted image region corresponds to a semantically meaningful and 
    species-discriminative part rather than background or noise.
    \item \textbf{Prototype Match:} whether the retrieved prototype 
    patches are visually similar to the highlighted region in the 
    test image.
    \item \textbf{Overall Adaptation Quality:} the overall semantic 
    convincingness of the model's prototype-based reasoning for 
    the predicted class.
\end{itemize}

Table~\ref{tab:vlm_scores} reports the complete results, including per-metric standard deviations. Furthermore, we measured the overall quality scores exclusively on samples where the models predicted correctly (mean$=4.62$ for ProtoTTA). These filtered scores consistently mirrored the general performance trends, demonstrating that the positive correlation between predictive accuracy and VLM-judged reasoning quality persists across adaptation methods.

\begin{table}[h]
    \centering
    \caption{Full VLM evaluation results on the 100-sample CUB-200-C subset. All scores on a 1--5 scale. Best in \textbf{bold}, second-best \underline{underlined}.}
    \label{tab:vlm_scores}
        \vspace{0.05in}
    \resizebox{\linewidth}{!}{%
        \begin{tabular}{l|ccc}
            \toprule
            \textbf{Method} 
            & \textbf{Focus Relevance} $\uparrow$ 
            & \textbf{Prototype Match} $\uparrow$ 
            & \textbf{Overall Quality} $\uparrow$ \\
            \midrule
            Unadapted  & 4.14 $\pm$ 0.98 & 3.66 $\pm$ 0.87 & 3.53 $\pm$ 1.02 \\
            Tent       & 4.03 $\pm$ 0.92 & 3.64 $\pm$ 0.92 & 3.59 $\pm$ 1.02 \\
            EATA       & \underline{4.18 $\pm$ 0.90} & \underline{3.79 $\pm$ 0.82} & \underline{3.75 $\pm$ 0.94} \\
            \rowcolor{gray!10}
            \textbf{\textit{ProtoTTA (Ours)}} & \textbf{4.30 $\pm$ 0.90} & \textbf{3.86 $\pm$ 0.81} & \textbf{3.78 $\pm$ 0.97} \\
            \bottomrule
        \end{tabular}%
    }
\end{table}

\subsection{PCA-W and VLM Correlation}
\label{sec:vlm_corr_analysis}

To validate that our proposed PCA-W metric captures genuine interpretability signals, we compute sample-level PCA-W from the saved prototype contribution scores and correlate them with the VLM's overall adaptation quality scores. Sample-level PCA-W is defined as:
\begin{equation}
    \mathrm{PCA\text{-}W}(x) = 
    \frac{\sum_{p \in \mathcal{P}_{GT}} c_p}
         {\sum_{p \in \mathcal{P}_{top}} c_p}
\end{equation}
where $\mathcal{P}_{top}$ are the top contributing prototypes for sample $x$, $\mathcal{P}_{GT} \subseteq \mathcal{P}_{top}$ are those belonging to the ground-truth class, and $c_p$ is the contribution score. A value of 1.0 means all top prototype mass comes from ground-truth class prototypes; 0.0 means none does.

Table~\ref{tab:vlm_corr} reports correlations pooled across all scored samples and within ProtoTTA alone. The moderate positive pooled correlation (Pearson $r{=}0.53$) confirms that PCA-W aligns with VLM-rated interpretability across diverse adaptation methods, establishing it as a valid, method-agnostic proxy for semantic reasoning.

Crucially, the correlation strengthens significantly when evaluated exclusively on ProtoTTA samples ($r{=}0.68$). This amplification highlights a fundamental mechanism of our approach: realigning mathematical activations with visual reality. Under distribution shifts, unadapted models and standard TTA baselines often suffer from ``semantic hallucinations''—mathematically assigning high activation weights to correct-class prototypes based on spurious matching with random background noise. This noise artificially inflates or deflates mathematical metrics, decoupling them from the actual visual evidence (which a VLM accurately scores as poor). By explicitly minimizing prototype activation entropy and employing geometric filtering, ProtoTTA suppresses these noise-induced spurious matches. Consequently, it ensures that when the mathematical score (PCA-W) is high, it is driven by genuinely discriminative, clean visual features, thus tightening the correlation with human-aligned VLM evaluations.

\begin{table}[h]
\centering
\caption{Pearson ($r$) and Spearman ($\rho$) correlations between sample-level PCA-W and VLM scores.}
\label{tab:vlm_corr}
\setlength{\tabcolsep}{5pt}
\small
\vspace{0.05in}
\begin{tabular}{l|cc|cc|cc}
    \toprule
    & \multicolumn{2}{c|}{\textbf{Overall Quality}} 
    & \multicolumn{2}{c|}{\textbf{Proto Match}} 
    & \multicolumn{2}{c}{\textbf{Focus Rel.}} \\
    \textbf{Subset} & $r$ & $\rho$ & $r$ & $\rho$ & $r$ & $\rho$ \\
    \midrule
    All ($N{=}397$)        & 0.53 & 0.59 & 0.49 & 0.56 & 0.32 & 0.39 \\
    ProtoTTA ($N{=}97$)    & 0.68 & 0.73 & 0.55 & 0.62 & 0.32 & 0.31 \\
    \bottomrule
\end{tabular}
\end{table}

\section{Explainable Test-Time Adaptation via Prototype Reasoning}
\label{sec:vlm_qual}
A key advantage of prototype-based architectures is that test-time adaptation is inherently explainable. Unlike black-box TTA methods that operate on output logits, prototype models expose the full evidence chain: which image regions are attended to, which training prototypes are retrieved, and how their contributions change under adaptation. This allows adaptation dynamics to be narrated in semantically meaningful terms.

\begin{figure}[h]
  \centering
  \includegraphics[width=\textwidth]{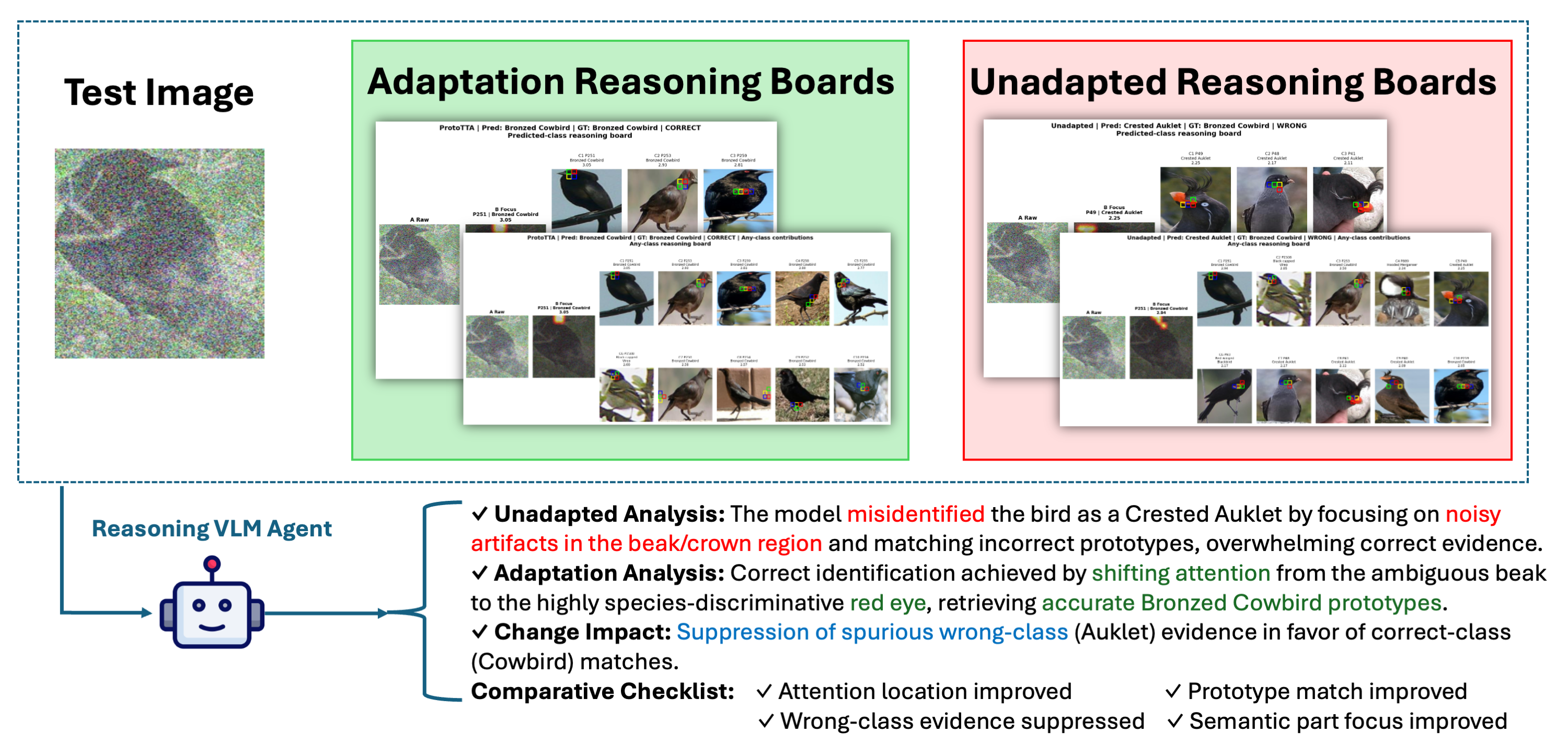}
  \caption{Explainable TTA via prototype reasoning boards. The 
  unadapted model (right) misclassifies the corrupted bird by 
  matching wrong-class prototypes in the noisy 
  beak/crown region. ProtoTTA (middle) corrects the prediction 
  by shifting to the species-discriminative red eye and retrieving correct-class Bronzed Cowbird prototypes, suppressing spurious 
  evidence. The VLM agent automatically narrates this before/after analysis from the reasoning boards.}
  \label{fig:vlm_qualitative}
\end{figure}

Figure~\ref{fig:vlm_qualitative} illustrates a representative 
example. The corrupted test image (left) shows a heavily noise-
corrupted bird. The unadapted model (right, red) misclassifies it as a Crested Auklet by fixating on noisy beak/crown artifacts — the reasoning board shows wrong-class Crested Auklet prototypes dominating both the predicted-class and any-class boards. After ProtoTTA adaptation (middle, green), the model corrects its prediction to the ground-truth Bronzed Cowbird. The reasoning boards reveal \textit{why}: the model shifts attention to the species-discriminative red eye region and retrieves multiple correct-class Bronzed Cowbird prototypes, while suppressing the spurious wrong-class evidence visible in the unadapted boards.

The VLM agent automatically generates this comparative analysis 
from the reasoning boards, producing structured output covering: unadapted failure analysis, adaptation success analysis, change impact, and a comparative checklist (attention location, wrong-class suppression, prototype match, semantic part focus). This framework enables language-grounded diagnosis of TTA behaviour that is uniquely possible with prototype-based models.

\section{ProtoTTA Pseudocode} \label{sec:peseudocode}
In this section, we provide the detailed pseudocode for our proposed method, ProtoTTA. Algorithm \ref{alg:prototta} outlines the complete adaptation procedure, including the geometric filtering mechanism and the prototype-guided entropy minimization steps.
\begin{algorithm}[h]
\caption{ProtoTTA: Test-Time Prototype Updates}
\label{alg:prototta}
\begin{algorithmic}[1]
\Require Pre-trained prototype model $f_\theta$, test stream $\mathcal{D}_{test}$, similarity threshold $\tau$, learning rate $\eta$
\Ensure Adapted parameters $\Theta$ (normalization layers + structural add-ons)
\State Initialize $\Theta \leftarrow \Theta_{init}$ \Comment{LayerNorm/BatchNorm + Attention Biases/1×1 Convs}
\For{each batch $\{\mathbf{x}_i\}_{i=1}^B \in \mathcal{D}_{test}$}
    \State \textcolor{gray}{// Forward pass and pseudo-labeling}
    \State Compute prototype similarities $s_{ip}$ for all prototypes $p$
    \State Map similarities to probability space: $\bar{s}_{ip} \in [0,1]$ \Comment{Linear scaling or sigmoid or log-scale}
    \State Obtain pseudo-labels: $\hat{y}_i \leftarrow \arg\max_c f_\theta(\mathbf{x}_i)$
    
    \State \textcolor{gray}{// Geometric filtering for reliable set}
    \State $\mathcal{R} \leftarrow \{ i \mid \max_p(\bar{s}_{ip}) > \tau \}$ \Comment{Optional: add entropy constraint}
    
    \If{$|\mathcal{R}| > 0$}
        \State \textcolor{gray}{// Consensus aggregation (Top-K Mean for sub-prototypes)}
        \State Aggregate sub-prototype scores via Top-K Mean
        
        \State \textcolor{gray}{// Compute weighted binary entropy loss}
        \State $\mathcal{L}_{ProtoTTA} \leftarrow \frac{1}{|\mathcal{R}|} \sum_{i \in \mathcal{R}} c_i \cdot \sum_{p \in \mathcal{P}_{\hat{y}_i}} w_p \cdot H(\bar{s}_{ip})$
        \State \quad where $H(\bar{s}_{ip}) = -\bar{s}_{ip} \log(\bar{s}_{ip}) - (1-\bar{s}_{ip}) \log(1-\bar{s}_{ip})$
        \State \quad $c_i$: confidence score, $w_p$: prototype importance weight, $\mathcal{P}_{\hat{y}_i}$: target prototypes
        
        \State \textcolor{gray}{// Update only normalization + structural parameters}
        \State $\Theta \leftarrow \Theta - \eta \nabla_{\Theta} \mathcal{L}_{ProtoTTA}$
    \EndIf
\EndFor
\State \Return Adapted model $f_{\Theta}$
\end{algorithmic}
\end{algorithm}

\section{Ablation Studies} \label{sec:ablation}
In this section, we provide a detailed analysis of the architectural and algorithmic choices governing \textit{ProtoTTA} for ProtoViT backbone. We validate the necessity of our geometric filtering, the selection of update parameters, and our consensus strategies.

\subsection{Impact of Geometric Filtering and Update Parameters}
Table \ref{tab:ablation_critical} illustrates the impact of our two most critical components: the Geometric Filter and the choice of trainable parameters.

\paragraph{Geometric Filtering.} The inclusion of the Geometric Filter is the most decisive factor in our framework. As shown in the first section of Table \ref{tab:ablation_critical}, removing the filter (adapting on all samples regardless of reliability) causes a drastic performance drop of nearly 4\% (60.12\% $\to$ 56.33\%). This confirms our hypothesis that adapting to ambiguous or low-confidence samples introduces noise that degrades the model, whereas restricting updates to the Reliable Set $\mathcal{R}$ ensures positive transfer.

\paragraph{Adaptation Parameters.} We investigate which parts of the model should be updated during test-time. Updating all learnable parameters (``All Adaptive'') leads to model collapse (51.86\%), likely due to catastrophic forgetting on the small batch size. Restricting updates to LayerNorm parameters stabilizes the model (59.41\%), a finding consistent with standard TTA literature. However, our proposed strategy of updating Attention Biases alongside LayerNorms yields the best performance (60.12\%). This suggests that adjusting the attention mechanism is crucial for prototype-based models to "re-focus" on relevant semantic features under corruption.

\begin{table}[h]
    \centering
    \caption{Comparison of Geometric Filtering Strategies and Adaptation Modes on ProtoViT. The ``Geometric Filter'' is essential for stability, preventing the model from learning on noisy data. For parameters, adding Attention Bias to LayerNorm updates yields the best results.}
    \label{tab:ablation_critical}
    \setlength{\tabcolsep}{10pt}
    \begin{tabular}{lcccc}
        \toprule
        \textbf{Configuration} & \textbf{Mean Acc} & \textbf{Std Dev} & \textbf{Min} & \textbf{Max} \\
        \midrule
        \multicolumn{5}{l}{\textit{\textbf{Geometric Filtering}}} \\
        No Filter & 56.33\% & 13.74 & 34.90 & 74.61 \\
        \rowcolor{gray!10} \textbf{With Filter (Ours)} & \textbf{60.12\%} & \textbf{10.55} & \textbf{43.29} & \textbf{74.96} \\
        \midrule
        \multicolumn{5}{l}{\textit{\textbf{Adaptation Parameters}}} \\
        All Adaptive & 51.86\% & 13.03 & 36.49 & 71.06 \\
        LayerNorm Only & 59.41\% & 10.89 & 42.04 & 75.06 \\
        \rowcolor{gray!10} \textbf{LN + Attn Bias (Ours)} & \textbf{60.12\%} & \textbf{10.55} & \textbf{43.29} & \textbf{74.96} \\
        \bottomrule
    \end{tabular}
\end{table}

\subsection{Design Refinements: Aggregation and Weighting}
Table \ref{tab:ablation_refinements} examines the finer design choices regarding prototype aggregation, target selection, and loss weighting.

\paragraph{Consensus Strategy.} Standard prototype models typically use the \texttt{max} operator to aggregate sub-prototype scores. While \texttt{max} performs well (60.04\%), it is sensitive to outliers. Our Top-K Mean strategy marginally outperforms it (60.12\%) and offers lower variance (10.55 Std Dev), validating that a consensus-based approach is more robust to the noise induced by corruptions.

\paragraph{Weighting and Target Selection.} We observe that weighting samples by both confidence ($c_i$) and prototype importance ($w_p$) yields the highest accuracy, though the gain over unweighted adaptation is incremental. Similarly, we compared adapting on ``Target Prototypes Only'' (derived from pseudo-labels) versus ``All Prototypes''. The results are nearly identical, justifying our design choice to use ``Target Only'' for its computational efficiency without sacrificing accuracy.

\begin{table}[h!]
    \centering
    \caption{Ablation of Consensus strategies, Prototype Targeting, and Loss Weighting on ProtoViT. Our choices (Top-K Mean, Target Only, and Combined Weighting) consistently provide the most robust performance.}
    \label{tab:ablation_refinements}
    \setlength{\tabcolsep}{8pt}
    \begin{tabular}{lcccc}
        \toprule
        \textbf{Configuration} & \textbf{Mean Acc} & \textbf{Std Dev} & \textbf{Min} & \textbf{Max} \\
        \midrule
        \multicolumn{5}{l}{\textit{\textbf{Consensus Strategy}}} \\
        Mean & 59.75\% & 10.77 & 42.68 & 75.22 \\
        Max (Standard) & 60.04\% & 10.59 & 43.22 & 74.89 \\
        \rowcolor{gray!10} \textbf{Top-K Mean (Ours)} & \textbf{60.12\%} & \textbf{10.55} & \textbf{43.29} & \textbf{74.96} \\
        \midrule
        \multicolumn{5}{l}{\textit{\textbf{Target vs. All Prototypes}}} \\
        All Prototypes & 60.11\% & 10.57 & 42.94 & 74.99 \\
        \rowcolor{gray!10} \textbf{Target Only (Ours)} & \textbf{60.12\%} & \textbf{10.55} & \textbf{43.29} & \textbf{74.96} \\
        \midrule
        \multicolumn{5}{l}{\textit{\textbf{Weighting Strategy}}} \\
        No Weighting & 60.03\% & 10.62 & 42.91 & 74.91 \\
        Importance Only & 60.04\% & 10.61 & 42.91 & 74.92 \\
        Confidence Only & 60.07\% & 10.65 & 42.79 & 75.03 \\
        \rowcolor{gray!10} \textbf{Both (Ours)} & \textbf{60.12\%} & \textbf{10.55} & \textbf{43.29} & \textbf{74.96} \\
        \bottomrule
    \end{tabular}
\end{table}

\section{Qualitative Analysis} \label{sec:qualitative_examples}
In this section, we leverage the inherent interpretability of prototype-based models to visually diagnose failure modes under distribution shifts and compare the effectiveness of different adaptation strategies. Unlike black-box models, this architecture allows us to trace the reasoning process by examining which prototypes are activated and how strongly they contribute to the final prediction.

\begin{figure}[h!]
  \centering
  \includegraphics[width=\textwidth]{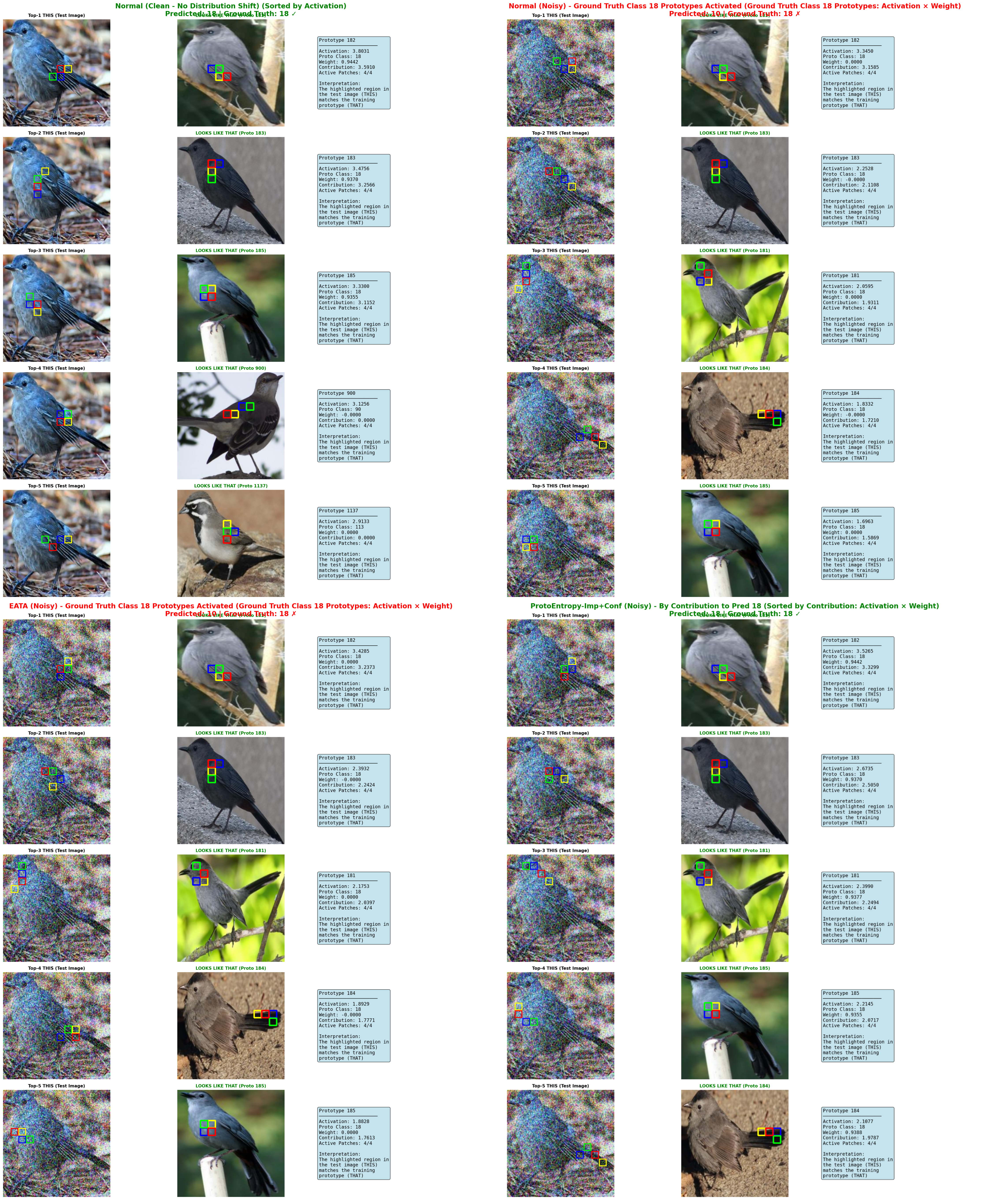}
  \caption{Visualization of Prototype Contributions. We compare the activation strengths of ground-truth class prototypes across three scenarios: (Top-Left) Normal model on clean data prototype activations (global), serving as the ``Gold Standard''; (Top-Right) Unadapted model on noisy data, showing suppressed activations; (Bottom-Left) EATA adaptation, which yields incomplete recovery; and (Bottom-Right) \textit{ProtoTTA}, which successfully restores the activation profile to match the clean baseline.}
  \label{fig:qual_analysis}
\end{figure}

Figure \ref{fig:qual_analysis} compares the activation intensity of the top-k prototypes across different scenarios on ProtoViT. The clean baseline (Top Left) displays the ideal global activation landscape, where the model attends to a mixture of relevant prototypes. Under corruption, the unadapted model (Top Right) exhibits a distinct failure mode: activations for the ground-truth class are severely suppressed. In contrast, prototypes belonging to incorrect classes are spuriously sharpened, causing the model to ``hallucinate'' and misclassify. While EATA (Bottom Left) attempts to adapt, it struggles to suppress these hallucinations or fully restore the signal strength of the correct class. In contrast, \textit{ProtoTTA} (Bottom Right) successfully recovers the original semantic focus. It restores the activation profile to closely mirror the clean baseline, thereby suppressing noise-induced hallucinations and enabling the model to make accurate predictions with high confidence.

\begin{figure}[h!]
  \centering
  \includegraphics[width=\textwidth]{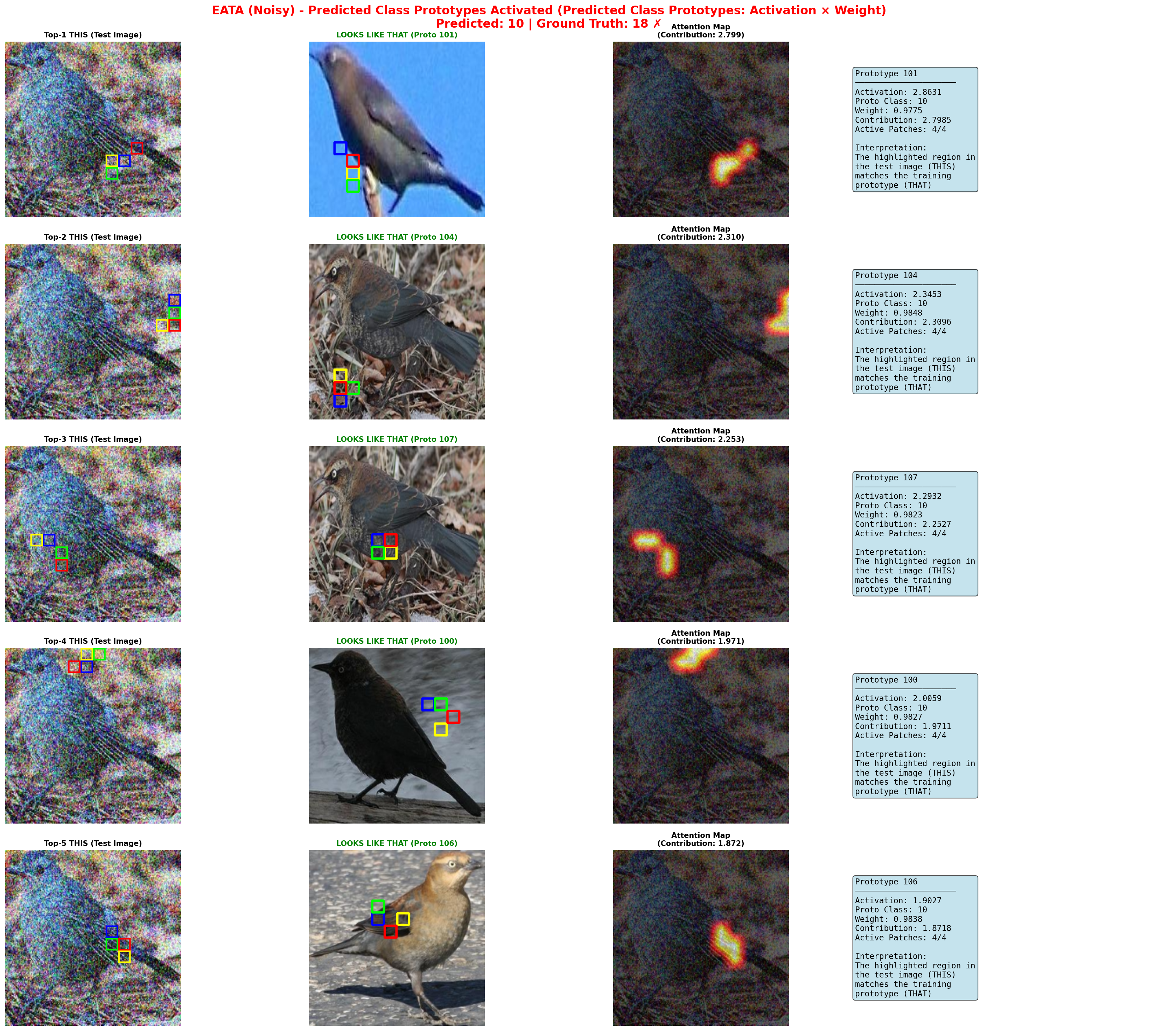}
  \caption{
  Analysis of EATA Misclassification. We visualize the top-5 activated prototypes for the incorrect class predicted by EATA. Despite the mismatch, the model registers high activation scores (hallucinations), indicating that the adaptation failed to filter out noise-induced features. This contrasts with Figure \ref{fig:qual_analysis}, where ground-truth activations were suppressed.
  }
  \label{fig:eata_failure}
\end{figure}

Figure \ref{fig:eata_failure} provides a deeper insight into why baseline methods fail. By visualizing the top-activated prototypes for the \textit{incorrect} class predicted by EATA, we observe that the model assigns high activation scores to irrelevant semantic features. This confirms that EATA fails to suppress noise-induced artifacts effectively. Instead, the noise deceives the model into ``hallucinating'' strong matches with incorrect prototypes. This phenomenon, amplified activations for the wrong class (Figure \ref{fig:eata_failure}) combined with suppressed activations for the correct class (Figure \ref{fig:qual_analysis}), fully explains the drop in robustness.

\begin{figure}[t!]
  \centering
  \includegraphics[width=\textwidth]{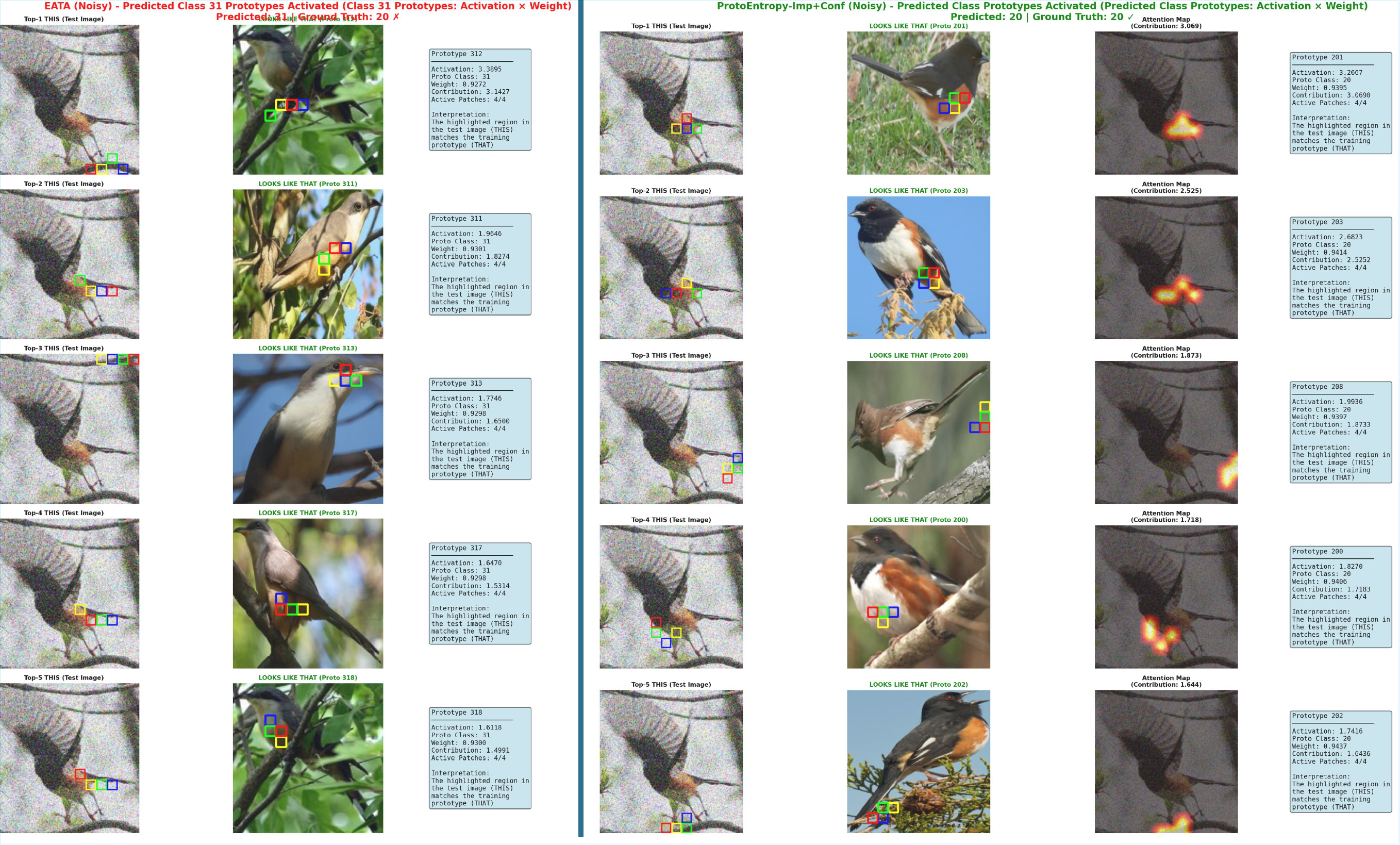}
  \caption{Comparison of Prototype Contributions (Activation $\times$ Weight) for the predicted class between EATA (left) and ProtoTTA (right) under noisy conditions. EATA suffers from semantic hallucination, confidently activating irrelevant prototypes for an incorrect class (Predicted: 31, Ground Truth: 20) based on spurious background artifacts. In contrast, ProtoTTA accurately maps spatial features to the ground-truth class prototypes (Predicted: 20, Ground Truth: 20), effectively suppressing noise-induced hallucinations and restoring correct semantic focus.}
  \label{fig:halluc}
\end{figure}

To further illustrate the contrast in adaptation dynamics, Figure \ref{fig:halluc} provides a direct comparison of the top-activated prototypes for the predicted class between EATA and ProtoTTA on a corrupted sample (Ground Truth Class: 20). While EATA incorrectly predicts Class 31 by hallucinating strong semantic matches with irrelevant features, ProtoTTA successfully suppresses these noise-induced artifacts. By explicitly minimizing prototype activation entropy , ProtoTTA recovers the correct semantic associations, attending to the appropriate visual features and accurately predicting the ground-truth class.

\begin{figure}[h!]
  \centering
  \includegraphics[width=\textwidth]{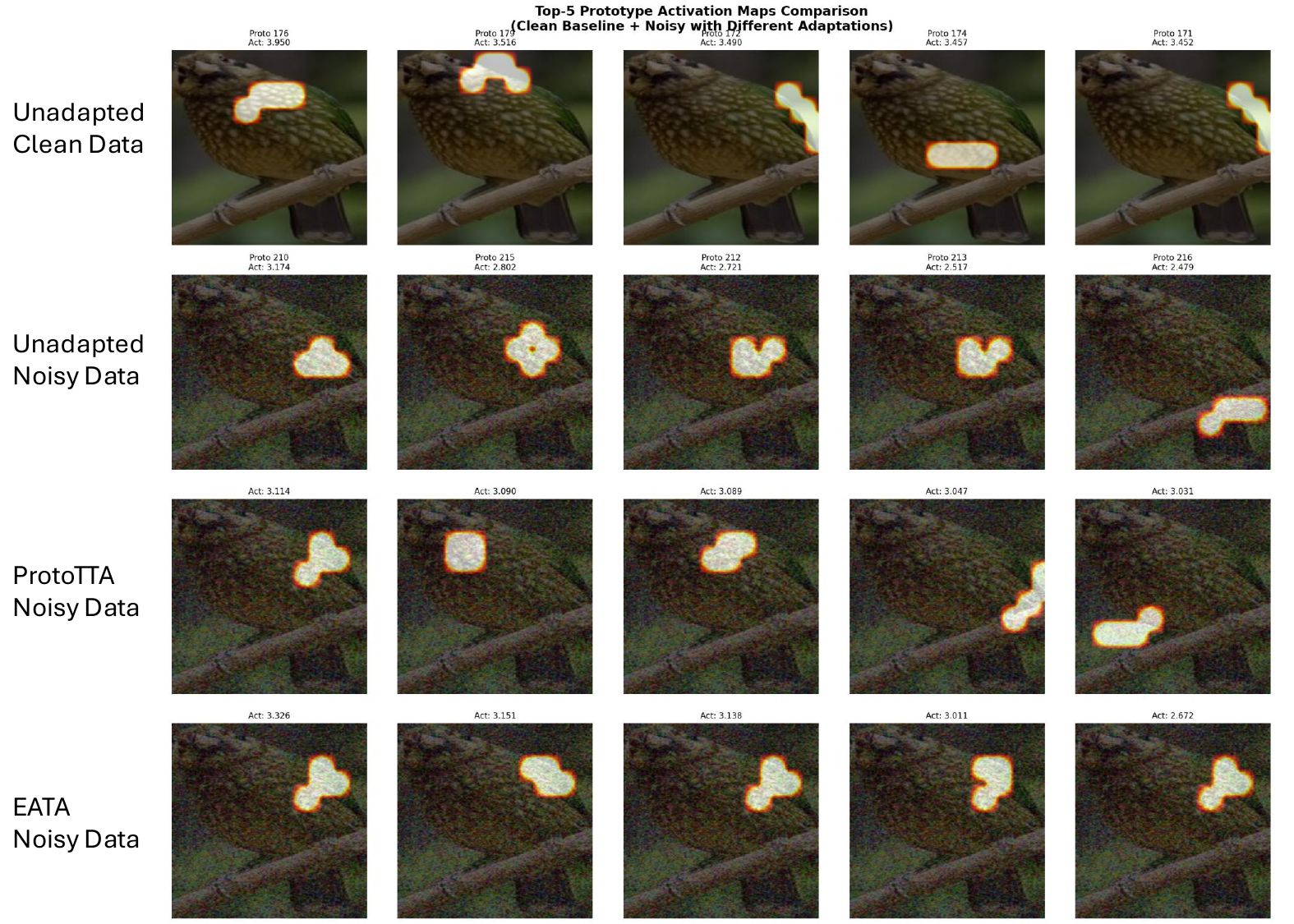}
\caption{Comparison of Prototype Attention Maps. We visualize the spatial regions of input samples triggering the top prototype activations across different settings. \textit{ProtoTTA} (Third Row) successfully realigns the model's attention, closely matching the focus of the Clean Baseline (Top Row) on the object of interest. Conversely, the Unadapted (Second Row) and EATA (Bottom Row) models diverge significantly, consistently attending to non-informative background regions or noise artifacts.}
  \label{fig:heatmap}
\end{figure}

Finally, Figure \ref{fig:heatmap} visualizes the spatial distribution of prototype activations. We observe that \textit{ProtoTTA} acts as a high-fidelity approximation of the uncorrupted model, triggering the same semantic regions of interest as the clean baseline. In contrast, both the Unadapted model and EATA (the strongest baseline) fail to attend to these discriminative features. Instead, they drift towards irrelevant background noise or spurious artifacts, likely ``easier'' high-frequency paths for the corrupted model to latch onto, which directly leads to the hallucinations and misclassifications discussed earlier.

\section{Forgetting Analysis} \label{sec:continual}
\begin{figure}[h!]
  \centering
  \includegraphics[width=\textwidth]{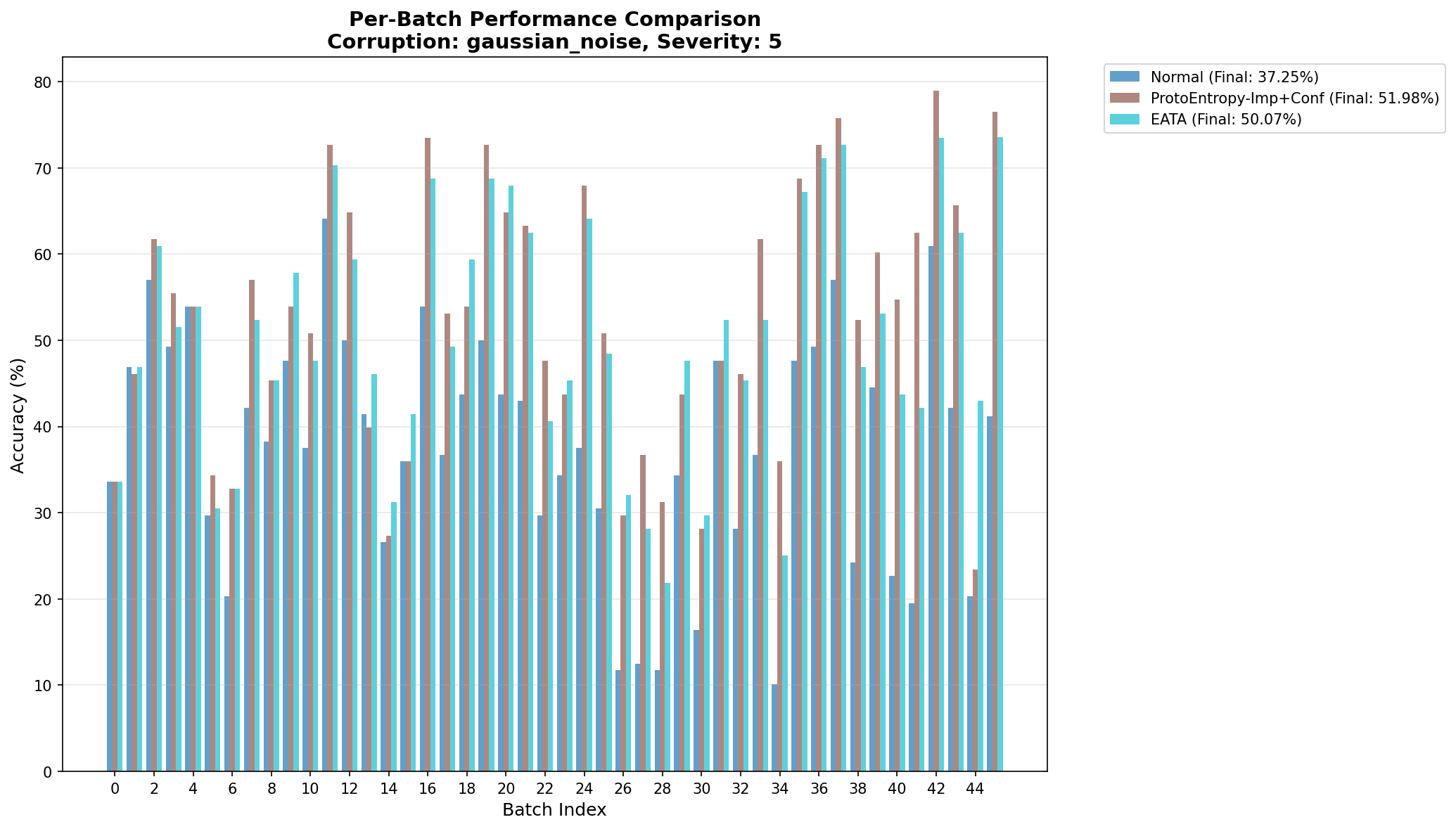}
  \caption{Per-Batch Performance Stability. We track classification accuracy across consecutive batches for Gaussian Noise (Severity 5). \textit{ProtoTTA} (Brown bars) consistently outperforms the Unadapted baseline (Blue) and remains competitive with EATA (Cyan) throughout the sequence. The absence of performance degradation in later batches confirms that \textit{ProtoTTA} avoids catastrophic forgetting, successfully maintaining stability by restricting updates to the reliable set.}
  \label{fig:forget}
\end{figure}

In our primary experimental setup, the model is reset between different corruption types. To investigate the stability of \textit{ProtoTTA} during prolonged adaptation on ProtoViT, we analyze the performance trend over a single long-sequence corruption (Gaussian Noise, Severity 5, approx. 5,500 samples) without intermediate resets. Figure \ref{fig:forget} illustrates the per-batch accuracy of \textit{ProtoTTA} compared to EATA and the unadapted baseline. We observe that \textit{ProtoTTA} maintains a consistent performance advantage over the unadapted model throughout the entire sequence. Crucially, there is no downward trend in accuracy across later batches, indicating that our method does not suffer from \textit{catastrophic forgetting} or error accumulation. The performance fluctuations align with the intrinsic difficulty of specific batches (mirrored by the unadapted baseline and EATA), confirming that our geometric filtering and reliability-weighted updates effectively prevent semantic drift.

\end{document}